\PassOptionsToPackage{unicode}{hyperref}
\PassOptionsToPackage{hyphens}{url}
\PassOptionsToPackage{dvipsnames,svgnames,x11names}{xcolor}
\documentclass[
]{article}
\usepackage{xcolor}
\usepackage{amsmath,amssymb}
\usepackage{iftex}
\ifPDFTeX
  \usepackage[T1]{fontenc}
  \usepackage[utf8]{inputenc}
  \usepackage{textcomp} 
\else 
  \usepackage{unicode-math} 
  \defaultfontfeatures{Scale=MatchLowercase}
  \defaultfontfeatures[\rmfamily]{Ligatures=TeX,Scale=1}
\fi
\usepackage{lmodern}
\ifPDFTeX\else
\fi
\IfFileExists{upquote.sty}{\usepackage{upquote}}{}
\IfFileExists{microtype.sty}{
  \usepackage[]{microtype}
  \UseMicrotypeSet[protrusion]{basicmath} 
}{}
\makeatletter
\@ifundefined{KOMAClassName}{
  \IfFileExists{parskip.sty}{%
    \usepackage{parskip}
  }{
    \setlength{\parindent}{0pt}
    \setlength{\parskip}{6pt plus 2pt minus 1pt}}
}{
  \KOMAoptions{parskip=half}}
\makeatother
\makeatletter
\ifx\paragraph\undefined\else
  \let\oldparagraph\paragraph
  \renewcommand{\paragraph}{
    \@ifstar
      \xxxParagraphStar
      \xxxParagraphNoStar
  }
  \newcommand{\xxxParagraphStar}[1]{\oldparagraph*{#1}\mbox{}}
  \newcommand{\xxxParagraphNoStar}[1]{\oldparagraph{#1}\mbox{}}
\fi
\ifx\subparagraph\undefined\else
  \let\oldsubparagraph\subparagraph
  \renewcommand{\subparagraph}{
    \@ifstar
      \xxxSubParagraphStar
      \xxxSubParagraphNoStar
  }
  \newcommand{\xxxSubParagraphStar}[1]{\oldsubparagraph*{#1}\mbox{}}
  \newcommand{\xxxSubParagraphNoStar}[1]{\oldsubparagraph{#1}\mbox{}}
\fi
\makeatother

\usepackage{color}
\usepackage{fancyvrb}

\DefineVerbatimEnvironment{Highlighting}{Verbatim}{commandchars=\\\{\}}
\usepackage{framed}
\definecolor{shadecolor}{RGB}{241,243,245}
\newenvironment{Shaded}{\begin{snugshade}}{\end{snugshade}}

\newcommand{\AttributeTok}[1]{\textcolor[rgb]{0.40,0.45,0.13}{#1}}

\newcommand{\CommentTok}[1]{\textcolor[rgb]{0.37,0.37,0.37}{#1}}

\newcommand{\ExtensionTok}[1]{\textcolor[rgb]{0.00,0.23,0.31}{#1}}

\newcommand{\NormalTok}[1]{\textcolor[rgb]{0.00,0.23,0.31}{#1}}

\usepackage{longtable,booktabs,array}
\usepackage{calc} 
\usepackage{etoolbox}
\makeatletter
\patchcmd\longtable{\par}{\if@noskipsec\mbox{}\fi\par}{}{}
\makeatother
\IfFileExists{footnotehyper.sty}{\usepackage{footnotehyper}}{\usepackage{footnote}}
\makesavenoteenv{longtable}
\usepackage{graphicx}
\makeatletter
\newsavebox\pandoc@box
\newcommand*\pandocbounded[1]{
  \sbox\pandoc@box{#1}%
  \Gscale@div\@tempa{\textheight}{\dimexpr\ht\pandoc@box+\dp\pandoc@box\relax}%
  \Gscale@div\@tempb{\linewidth}{\wd\pandoc@box}%
  \ifdim\@tempb\p@<\@tempa\p@\let\@tempa\@tempb\fi
  \ifdim\@tempa\p@<\p@\scalebox{\@tempa}{\usebox\pandoc@box}%
  \else\usebox{\pandoc@box}%
  \fi%
}
\def\fps@figure{htbp}
\makeatother

\NewDocumentCommand\citeproctext{}{}

\makeatletter
 \let\@cite@ofmt\@firstofone
 \def\@biblabel#1{}
 \def\@cite#1#2{{#1\if@tempswa , #2\fi}}
\makeatother
\newlength{\cslhangindent}
\newlength{\csllabelwidth}
\newenvironment{CSLReferences}[2] 
 {\begin{list}{}{%
  \setlength{\itemindent}{0pt}
  \setlength{\leftmargin}{0pt}
  \setlength{\parsep}{0pt}
  \ifodd #1
   \setlength{\leftmargin}{\cslhangindent}
   \setlength{\itemindent}{-1\cslhangindent}
  \fi
  \setlength{\itemsep}{#2\baselineskip}}}
 {\end{list}}
\usepackage{calc}

\providecommand{\tightlist}{%
  \setlength{\itemsep}{0pt}\setlength{\parskip}{0pt}}

\usepackage{arxiv}
\usepackage{orcidlink}
\usepackage{amsmath}
\usepackage[T1]{fontenc}
\makeatletter
\@ifpackageloaded{caption}{}{\usepackage{caption}}
\AtBeginDocument{%
\ifdefined\contentsname
  \renewcommand*\contentsname{Table of contents}
\else
  \newcommand\contentsname{Table of contents}
\fi
\ifdefined\listfigurename
  \renewcommand*\listfigurename{List of Figures}
\else
  \newcommand\listfigurename{List of Figures}
\fi
\ifdefined\listtablename
  \renewcommand*\listtablename{List of Tables}
\else
  \newcommand\listtablename{List of Tables}
\fi
\ifdefined\figurename
  \renewcommand*\figurename{Figure}
\else
  \newcommand\figurename{Figure}
\fi
\ifdefined\tablename
  \renewcommand*\tablename{Table}
\else
  \newcommand\tablename{Table}
\fi
}
\@ifpackageloaded{float}{}{\usepackage{float}}
\floatstyle{ruled}
\@ifundefined{c@chapter}{\newfloat{codelisting}{h}{lop}}{\newfloat{codelisting}{h}{lop}[chapter]}
\floatname{codelisting}{Listing}

\makeatother
\makeatletter
\@ifpackageloaded{caption}{}{\usepackage{caption}}
\@ifpackageloaded{subcaption}{}{\usepackage{subcaption}}
\makeatother
\usepackage{bookmark}
\IfFileExists{xurl.sty}{\usepackage{xurl}}{} 
\hypersetup{
  pdftitle={Building Models of Neurological Language},
  colorlinks=true,
  linkcolor={blue},
  filecolor={Maroon},
  citecolor={Blue},
  urlcolor={Blue},
  pdfcreator={LaTeX via pandoc}}

\newcommand{\runninghead}{A Preprint }
\title{Building Models of Neurological Language}
\author{\textbf{Henry Watkins}~\orcidlink{0000-0001-6330-6195}\\\\UCL
Queen Square Institute of
Neurology\\\\\href{mailto:h.watkins@ucl.ac.uk}{h.watkins@ucl.ac.uk}}
\date{}
\begin{document}
\maketitle
\begin{abstract}
This report documents the development and evaluation of domain-specific
language models for neurology. Initially focused on building a bespoke
model, the project adapted to rapid advances in open-source and
commercial medical LLMs, shifting toward leveraging retrieval-augmented
generation (RAG) and representational models for secure, local
deployment. Key contributions include the creation of neurology-specific
datasets (case reports, QA sets, textbook-derived data), tools for
multi-word expression extraction, and graph-based analyses of medical
terminology. The project also produced scripts and Docker containers for
local hosting. Performance metrics and graph community results are
reported, with future possible work open for multimodal models using
open-source architectures like phi-4.
\end{abstract}

\section{Introduction}\label{sec-introduction}

This project was initially proposed to deal with the scale and
complexity of neurological text as used in practice. With a generous
fellowship provided by the Guarantors of Brain, I planned a multistage
project to tackle the problem of text in medical neuroscience using
recent advances in natural language processing. Neurological information
is still recorded almost entirely in free-form text---through clinic
letters, inpatient notes and investigative reports. This unstructured
approach reflects the vast complexity of the nervous system: not only
the signs and symptoms we observe in patients but also the detailed
anatomy, physiology and pathology revealed by various tests. That
complexity creates a Catch-22 for anyone hoping to build accurate,
machine-readable models of neurological data. On the one hand, you need
richly detailed, structured records to capture the full subtleties of
real-world practice. On the other hand, you can only analyse and
structure data at scale by relying on automated methods---methods that
won't work without those very structured records in the first place. In
practice, most projects resolve this dilemma in one of two ways:

\begin{enumerate}
\def\labelenumi{\arabic{enumi}.}
\tightlist
\item
  \textbf{Oversimplification}: They flatten complex neurological
  findings into a narrow set of standardised fields, which limits how
  faithfully the data reflect true clinical nuance.
\item
  \textbf{Manual coding}: They painstakingly annotate and parameterise
  free-text records by hand, which preserves detail but is too slow and
  labour-intensive to apply on a large scale.
\end{enumerate}

Arguably, the only way to break this deadlock is with a machine model
grounded in logical grammar. By precisely defining the facts we care
about and their interrelationships, such a model could automatically
extract structured information from the naturally occurring language of
clinical records. Traditional NLP methods aim for this goal but fall
short: their limited expressivity compromises both factual accuracy and
the ability to capture complex relations, and they remain technically
demanding to build and maintain---a reality reflected by their scant
adoption in healthcare settings (Topol 2019). What we urgently need is a
new generation of NLP tools with three essential features:

\begin{enumerate}
\def\labelenumi{\arabic{enumi}.}
\tightlist
\item
  \textbf{Rich logico-grammatical expressivity}, to represent intricate
  medical facts and relationships faithfully.
\item
  \textbf{Non-technical operability}, so neurologists and
  neuroscientists can design and run tasks without any programming.
\item
  \textbf{Safe clinical deployment}, ensuring seamless integration into
  patient-care environments.
\end{enumerate}

This kind of NLP would generate high-fidelity, structured descriptions
of neurological phenomena and enable domain experts to work directly
with their data---finally resolving the Catch-22 that has stalled
progress toward truly nuanced, individualised models in neurology. In
this project, I set out to apply the latest advances in natural language
modelling specifically to neurological texts. As part of my Brain
Non-Clinical Fellowship, my goals were twofold: first, to advance our
understanding of how language is used in clinical neurology; and second,
to develop practical resources or tools that support that work,
delivering research and outputs with both scientific and clinical value.

\subsection{Initial Goals}\label{initial-goals}

The initial aim of this project was to build Neurobase, a locally
deployable foundational language model tailored for neurology, by
adapting and extending state-of-the-art large language models (LLMs)
(Brown et al. 2020). Neurobase was envisioned as a system that would
combine the logical-grammatical strengths of modern LLMs (themselves
subject to some problems (Omiye et al. 2024)) with deep, domain-specific
knowledge extracted from thousands of unstructured inpatient and
outpatient records at the National Hospital for Neurology and
Neurosurgery (NHNN).

The core idea was to enable clinicians and researchers---regardless of
programming expertise---to interact with neurological data using natural
language instructions. For example, users could issue queries like
``Extract all mentions of tremor onset and progression from this set of
clinic letters,'' and Neurobase would return structured, relevant
information. This approach would go beyond simple information retrieval:
by leveraging the model's internal representations, Neurobase would be
able to generate detailed, hierarchical ``deep phenotypes'' that capture
both individual clinical features and the complex relationships between
them.

A further goal was to prototype a multimodal foundation model capable of
jointly processing free text and brain imaging data, thereby
demonstrating the added value of integrating these complementary
sources. Importantly, Neurobase was designed to run entirely on local
hardware, using its LLM backbone for logical-grammatical processing
rather than as an external ``knowledge oracle.'' This architecture was
intended to mitigate confidentiality and hallucination risks associated
with cloud-based models like GPT-3 or ChatGPT. The project was
structured around five main objectives:

\begin{enumerate}
\def\labelenumi{\arabic{enumi}.}
\tightlist
\item
  Fine-tune the best available LLM on large-scale, comprehensive
  free-text data from NHNN to create a foundational model of
  neurological language.
\item
  Develop a suite of natural language `prompt' queries for extracting
  structured phenotypic features from clinical and investigation text
  records.
\item
  Analyse the model's latent space to uncover a hierarchically
  organised, deep representation of neurological phenomena.
\item
  Validate the model and associated tools against manual
  parameterizations, both prospectively and externally using data from
  King's College and Guy's \& St Thomas's Hospitals.
\item
  Prototype a foundational model that integrates free text and brain
  imaging data.
\item
  Conduct a rigorous safety analysis to ensure alignment, privacy, and
  resistance to misuse.
\end{enumerate}

\subsection{Change of Focus}\label{change-of-focus}

Initially, our objective was to develop a fine-tuned, domain-specific
language model tailored to neurological practice. However, the pace of
innovation in large language models---particularly those adapted for
medical contexts---has been so rapid that by the time our bespoke model
would have been ready for deployment, its comparative advantage would
already have been eroded. There are a number of LLM-based systems
currently in use in general medical systems (Z. Wang et al. 2025),
including both open-source, academic projects (Moor et al. 2023) like
Apollo (X. Wang et al. 2024), GatorTron (Yang et al. 2022), MediTron
(Chen et al. 2023) and Almanac (Zakka et al. 2024); as well as models
from commercial providers such as Gemini (Saab et al. 2024; Team et al.
2024) and multimodal architectures (Wu et al. 2023).

In response, we redirected our efforts toward alternative architectures
and methodologies. One avenue has been the exploration of graph-based
language representations, which capture entities and their
interrelationships in a structured network; another has been the
adoption of retrieval-augmented generation (RAG) frameworks (Lewis et
al. 2020), wherein a powerful, general-purpose LLM is coupled with an
external knowledge base to ground its outputs in up-to-date,
domain-specific information.

Meanwhile, a host of ``medical copilots'' and foundation models have
emerged as commercial and open-source offerings, delivering high levels
of accuracy and usability out of the box. Despite our initial ambition
to produce an in-house model uniquely optimised for our data, empirical
evaluations have shown that these large, pre-trained
architectures---when supplemented by targeted retrieval
mechanisms---often match or exceed the performance of smaller, narrowly
fine-tuned systems (Soudani, Kanoulas, and Hasibi 2024; Nori et al.
2023). Example copilot-like medical model products include the following
list:

\begin{itemize}
\tightlist
\item
  \href{https://imerit.net/radiology-image-annotation-co-pilot/}{Medical
  Image Data Annotation Co-pilot iMerit}
\item
  \href{https://www.gleamer.ai/}{Gleamer Homepage AI for radiology -
  Gleamer}
\item
  \href{https://www.medicai.io/solutions/radiology-ai-co-pilot}{Efficient
  Radiology Workflows with AI Co-Pilots}
\item
  \href{https://medicalcopilot.eu/}{Medical Copilot - Customised \&
  Transparent LLMs for Healthcare}
\item
  \href{https://www.medsightai.com/}{MedSightAI : Spend Less Time
  Charting, More Time with Patients}
\item
  \href{https://techcommunity.microsoft.com/blog/healthcareandlifesciencesblog/a-deeper-look-at-microsoft-dragon-copilot-transforming-clinical-workflow-with-ai/4389496}{A
  Deeper Look at Microsoft Dragon Copilot: Transforming Clinical
  Workflow with AI \textbar{} Microsoft Community Hub}
\item
  \href{https://www.nature.com/articles/s41467-025-57426-0}{Towards a
  holistic framework for multimodal LLM in 3D brain CT radiology report
  generation \textbar{} Nature Communications}
\end{itemize}

This convergence of evidence leads to a clear conclusion: rather than
investing substantial resources in refining a standalone,
domain-specific LLM, it is more effective to harness a state-of-the-art
general model in tandem with a curated medical database (via RAG or
similar techniques). Such a hybrid approach yields both the broad
linguistic competence of large-scale pre-training and the precise,
factual grounding required for clinical applications.

The remainder of the report will detail the progress made on these new
approaches, and how they are implemented. These sub-projects consisted
of data curation (Section~\ref{sec-datasets}), vocabulary generation
(Section~\ref{sec-vocab}), deployment architecture
(Section~\ref{sec-deployment}), graphical term analysis
(Section~\ref{sec-termgraph}), phenotyping
(Section~\ref{sec-phenotypes}), model training (Section~\ref{sec-model})
and alignment (Section~\ref{sec-alignment}). These sections will also
document the artefacts produced thus far for each sub-project, such as
code bases, datasets and documents (Section~\ref{sec-artefacts}).

\section{Datasets}\label{sec-datasets}

In the course of this project, a significant focus has been on the
production of valuable datasets for training and evaluation. These have
been collected from neurology-specific sources from a range of external
text datasets alongside our own text corpora. The quality, format and
domain of training and test data are paramount in the creation of
language models, and work has shown that high performance can be
achieved with smaller training sets provided the quality is high and
selected for breadth (Gunasekar et al. 2023). The first stage of
building domain-specific models therefore is the collection of
high-quality text datasets tailored to the needs and data distribution
that will be encountered in practice. The approach I aimed for was a
mixture of local data produced by the department, web-scale data from
medical sources and multiple-choice-question-answering (MCQA) data.
Details about the location and accessing these datasets can be found in
Section~\ref{sec-artefacts}

\subsection{Radiology Reports}\label{radiology-reports}

The first data source was collected for a previous project involving
large-scale radiological reporting from the National Hospital for
Neurology and Neurosurgery (NHNN). These neuroradiological reports are
produced by radiologists for the task of diagnosis and monitoring. They
represent the gold standard of clinical-grade text for neurological and
radiological presentations. The dataset for this work is a corpus of
336,569 anonymised radiological reports extracted from the PACS system
of the National Hospital for Neurology and Neurosurgery (NHNN), London,
UK, for the purpose of local service evaluation and optimisation. This
corpus represents scans for a total of 181,519 unique patients from
August 1999 until February 2023. Each is a short narrative of between a
few dozen to a few hundred words describing a patient's condition and
scans. From these, we can identify the diagnosis and neurological health
status. From the point of view of a foundation model, these data capture
the idiosyncratic style of neurology alongside key vocabulary as it is
actually used in practice. More information on this dataset can be found
in our previous publication (Watkins et al. 2025). An example radiology
report is as follows: \textgreater{} There is moderate dilatation of the
third and lateral ventricles, distension of the third ventricular
recesses and mild enlargement of the pituitary fossa with depression of
the glandular tissue. Appearances are in keeping with hydrocephalus and
the mild associated frontal and peritrigonal transependymal oedema
indicates an ongoing active element. No cause for hydrocephalus is
demonstrated and the fourth ventricle and aqueduct are normal in
appearance. Note is made of T1 shortening within the floor of the third
ventricle. On the T2 gradient echo there is a bulbous focus of reduced
signal in this region that is inseparable from the terminal basilar
artery but likely to be lying anterosuperiorly. The combination of
features is most likely to reflect a small incidental dermoid within the
third ventricle with fat and calcified components. This could be
confirmed by examination of the CT mentioned on the request form. If the
presence of fat and calcium is not corroborated on the plain CT, a
time-of-flight MR sequence or CTA would be prudent to exclude the small
possibility of a vascular abnormality in this region.

\subsection{Specialist Examinations}\label{specialist-examinations}

Speciality Certificate Examinations are a necessary part of a
specialist's education in the UK medical system. These questions take
the form of complex multiple-choice exam questions above a difficult
patient presentation. Our dataset takes the corpus of
neurology-specialist questions from the Royal College of Physicians
\href{https://www.thefederation.uk/examinations/specialty-certificate-examinations/specialties/neurology}{website}
to build an MCQA dataset to evaluate the performance of our models. This
is a small collection (N=81) of multiple-choice questions sourced from
the MRCPUK neurology speciality exam. As the speciality exam, these
questions represent not just graduate-level text for clinical staff
found in current open-source datasets such as MMLU, but include
questions of greater difficulty and domain specificity than is available
in standard graduate medical texts. Furthermore, these questions are
expressed as diagnostic vignettes, whereby a patient's presentation is
described, and it is the task of the answerer to determine the most
likely diagnosis or course of treatment. Therefore these questions much
more closely match the idiom of clinical medicine than other available
medical datasets. As a small dataset, this set is used only for
evaluation, and no training or fine-tuning is performed with it.

In addition to this dataset, I note there are also other datasets of
medical MCQA questions available, such as the MMLU dataset (Y. Wang et
al. 2024) and the MedMCQA dataset (Pal, Umapathi, and Sankarasubbu
2022), the BioASQ dataset (Krithara et al. 2023) and the medical exam QA
dataset (Jin et al. 2021). These datasets are not used in this work but
may be useful for future work.

\subsection{NICE Guidelines}\label{nice-guidelines}

The National Institute for Health and Care Excellence is part of the UK
Department of Health and Social Care and is responsible for judging the
cost-effectiveness of medicines and providing guidance on clinical
practice. The
\href{https://www.nice.org.uk/guidance/conditions-and-diseases}{NICE
website} contains a large number of clinical guidelines, which are used
to inform clinical practice in the UK. These guidelines are written in a
structured format and contain information on the diagnosis, treatment
and management of various conditions. The NICE guidelines are an ideal
source of domain-specific text for training and evaluation of language
models, as they contain a wealth of information on the latest clinical
practice and are written in a clear and concise manner. The NICE
guidelines are available in PDF format and can be processed using GROBID
(Lopez 2009) to extract the structured text. The extracted text is then
converted into JSON format for downstream use. These guidelines are used
as a source of retrieval-augmented generation (RAG) data, where the
model is provided with additional context from the guidelines during
inference. This allows the model to generate more accurate and relevant
responses to queries related to NICE clinical practice.

\subsection{Case Report Datasets}\label{case-report-datasets}

Medical case reports are detailed, narrative accounts that describe
individual patient cases. They typically include information on the
patient's history, clinical presentation, diagnostic workups, treatment
interventions, and follow‐up results. These reports are valuable in
highlighting rare or novel conditions, serving as a basis for medical
education and further research. In this way, they can be considered the
`raw material' of medical research. This dataset has been constructed
out of a large corpus of case reports sourced via the UCL library
digital service. Case reports are an ideal source of neurological raw
text because they describe novel appearances beyond textbook
descriptions. By studying not just clean, textbook expositions of
diseases, conditions and pathological appearances, we can cover the
space of neurology as it appears in practice in a much more
comprehensive manner. The aim of this case report dataset is for the
foundation model to learn not just the archetypal presentations of
disease found in textbooks, but to learn reasoning across varied and
novel presentations of disease. The authors describe the patient's
history presentation, the investigations performed and finally, the
correct diagnosis, such that future readers will find useful when
encountering the same obscure presentation. We believe they form an
interesting source of clinical information since they represent the
diagnostic paradigm of encounter, description, investigation, diagnosis
and perhaps treatment that represents the normal working process of
medicine. In terms of language modelling, it also presents a new source
of difficult reasoning tasks that by their novel nature, would not be
present in the training set of conventional language models, medical or
otherwise.

The case reports in this work have been sourced from
case-report-specific journals such as the Journal of Medical Case
Reports, BMJ Case Reports. The primary use of these reports was as RAG
and pre-training data, but I have also created a text summarisation
dataset from these reports, where the task is to summarise the case
report into a short sentence containing the pertinent diagnosis. This is
a useful task for evaluating the performance of language models on
summarisation tasks and is a common task in the field of natural
language processing. Therefore I have two major datasets from this
source, one for pre-training and one for summarisation.

\subsection{Textbook Datasets}\label{textbook-datasets}

Medical textbooks and reference resources are the ideal standard of
domain-specific text. In theory, containing all the key information
necessary for a practitioner in a particular field, these text corpora
are the ideal source of gold-standard information for training or
testing language models. The textbooks used as source data for this
project are a collection of neurology-specific textbooks, which have
been processed to extract the relevant text using GROBID. The text comes
from (Howard et al. 2024; Sealfon, Stacy, and Motiwala 2016; Clarke
2022; Adams et al. 1997; Simon et al. 2009).

In addition to training data, we can also construct
multiple-choice-question-answering (MCQA) datasets from these textbooks,
which can be used to evaluate the performance of language models on
domain-specific tasks or fine-tune language models. To generate
multiple-choice questions from textbook data, we follow the procedure
outlined in Raina and Gales (2022), Mulla and Gharpure (2023) and Zhang
(2022). First, we choose a random passage from the textbook and generate
a question-answer pair using OpenAI's GPT-4. The false `distractor'
answers are then generated by a second call, such that the answers are
incorrect. In addition to human-author questions present in the text,
this produces a full dataset of MCQA questions that can be used to
evaluate the performance of language models on domain-specific tasks.
The questions are designed to be challenging and require a deep
understanding of the subject matter, making them suitable for evaluating
the performance of language models on complex tasks.

An example question from the textbook dataset is as follows:

\begin{quote}
question:``A 35-year-old man with HIV infection presents with complaints
of salivary gland enlargement, xerostomia, and uveitis. What syndrome
should be considered in the differential diagnosis?'', answers: (1)
HIV-associated neuropathy, (2)Sjögren's syndrome, (3) Demyelinating
peripheral neuropathy, (4) Diffuse inflammatory lymphocytosis syndrome
(DILS)
\end{quote}

\section{Neurological Multi-Word Expressions}\label{sec-vocab}

Multi-word expressions (MWEs) are lexical units composed of two or more
words whose meaning cannot always be derived from their individual
components (Villavicencio and Idiart 2019). In specialised domains MWEs
like \emph{small vessel ischaemia} or \emph{left frontal lesion} carry
domain‐specific semantics that is critical for tasks ranging from
information retrieval to knowledge‐base construction. As such, a key
task in developing a foundational model of neurological language is to
identify the phrases and expressions of interest in a data-driven way
that is unaffected by a-priori definitions of domain linguistic
structure. Domain‐specific corpora often contain frequent, semantically
rich phrases that cannot be captured by unigram models alone.
Identifying these MWEs is a crucial first step toward building robust
domain vocabularies and downstream NLP applications such as named‐entity
recognition, topic modelling, and term‐normalisation. For this purpose,
I created pyMWE, a Python toolkit, which (1) automatically extracts
candidate MWEs from raw text using statistical measures (specifically,
Pointwise Mutual Information), and (2) computes feature‐cluster
associations via Matthews Correlation Coefficients.

\subsection{Methodology}\label{methodology}

MWEs are traditionally extracted using co-occurrence frequency counts
and Positive Pointwise Mutual Information. \[
\mathrm{PPMI}(x, y) = \max\left(0, \log_2 \frac{p(x, y)}{p(x)\,p(y)}\right)
\]

It's a measure used in natural language processing and information
theory to quantify how much more often two events (like words) occur
together than if they were independent. p(x, y) is the probability that
both words x and y occur together (joint probability). p(x) is the
probability that word x occurs and p(y) is the probability that word y
occurs. The max function ensures that the value is non-negative, meaning
it only counts cases where the words occur together more often than
expected by chance. A higher PPMI value indicates a stronger association
between the two words, suggesting they are likely to form a meaningful
phrase or MWE. The value of using an information-theoretic measure like
PMI is it allows interpretability in a way lacking for
language-model-based measures or deep-learning metrics derived from
high-dimensional word vectors. Domain‐adapted MWE extraction often
requires configurable frequency thresholds to avoid noise in specialised
vocabularies (Church and Hanks 1990). The approach can be defined thus:

\begin{itemize}
\tightlist
\item
  The input is presented as a list of document strings, stop words are
  then removed. Currently, this uses a generic English list;
  domain‐specific stop‐words may improve quality.
\item
  Tokenising and counting n-grams (unigrams and bigrams) across
  documents. Future extensions could consider trigrams or skip-grams for
  richer MWEs.
\item
  The tokenised n-gram texts are vectorised.
\item
  Estimating bigram importance via Pointwise Mutual Information (PMI).
\item
  Let \(X_{\text{uni}}[i]\) be the total count of unigram \(i\) across
  corpora; \(X_{\text{bi}}[j]\) the count of bigram \(j\).
\item
  Estimate probabilities \(p(x) = X_{\text{uni}}[i]/N\) and
  \(p(x,y) = X_{\text{bi}}[j]/N\), where
  \(N = \sum_i X_{\text{uni}}[i]\).
\item
  Compute \(\mathrm{PMI}(x,y) = \log_2\bigl(p(x,y)/(p(x)p(y))\bigr)\),
  guarding against zero‐division.
\item
  Providing user‐configurable thresholds (document‐frequency bounds,
  number of MWEs). Here I also incorporate significance testing (e.g.,
  permutation tests) to filter spurious MWEs. I sort bigrams by
  descending PMI and return the top \texttt{n} as candidate MWEs.
\end{itemize}

In addition to identifying MWEs, the software also has tools for
feature-cluster correlation. This workflow allows rapid identification
of domain terms most characteristic of particular document clusters. If
one has a set of labels for a set of documents, we can identify in a
principled way the most important expressions that characterise each
particular label. The key statistic in this approach is \textbf{Matthews
Correlation Coefficient (MCC)}: a balanced measure for binary
classification quality, robust to class imbalance, and interpretable as
a correlation coefficient between observed and predicted binary labels.
It takes into account true positives (TP), true negatives (TN), false
positives (FP), and false negatives (FN), providing a single score that
balances all these outcomes. MCC is defined as:

\[
\mathrm{MCC} = \frac{TP \cdot TN - FP \cdot FN}{\sqrt{(TP + FP)(TP + FN)(TN + FP)(TN + FN)}}
\]

In this context, the MCC is used to quantify the association between
binary cluster membership (e.g., whether a document belongs to a
particular cluster) and the presence of specific MWEs. The higher the
MCC, the stronger the association between the MWE and the cluster.

\subsection{Applications}\label{applications}

In summary, pymwe allows me to identify the relevant MWEs in a corpus of
texts, and to correlate these MWEs with clusters of documents. These are
critical for understanding the language used in clinical documentation
or research literature but are often missed by simple word-based models.
Linking MWEs with document clusters (e.g., diagnosis categories) can
identify terms that are especially informative for each cluster, thus
providing a principled way to extract domain-specific vocabulary. This
is particularly useful in medical contexts, where precise terminology is
crucial for accurate communication and understanding. This leads to the
data-driven discovery of terminology that traditional glossaries miss.
MWEs serve as robust features for downstream tasks like classification
or concept normalisation and the information-theoretic measures or PPMI
and MCC offer transparent justifications for selected terms.

Imagine a dataset of radiology reports or narratives labelled by
diagnosis clusters (e.g., ``stroke'', ``dementia'', ``tumour'').

Extracted MWE: periventricular white matter changes

Cluster Association: Strong MCC with ``dementia'' cluster

Use: This MWE can now serve as a feature in a classifier predicting
dementia-related conditions, or as a token in a domain-adapted language
model fine-tuned on neurological data.

Links to the software and its documentation can be found in
Section~\ref{sec-artefacts}

\section{Modelling Neurological Language}\label{sec-model}

Neurology remains primarily expressed in unstructured language---clinic
letters, inpatient records, and investigational reports---rendering it
challenging to harness data at scale. Our goal is to develop a locally
deployable language model (LLM), \emph{NeuroBase}, that matches or
exceeds the performance of large benchmark models (e.g., GPT-4) on
neurology-specific tasks while maintaining data security and minimising
computational overhead. To achieve this, we fine-tune small open-source
Google Gemma models (Team et al. 2024) under two conditions: with and
without retrieval-augmented generation (RAG) (Lewis et al. 2020). We
evaluate these configurations against two newly introduced neurology
datasets and compare results to GPT-4. Unlike copilot-style systems,
which have grown ubiquitous in general radiology, our modular approach
leverages existing radiology-specific segmentation, classification, and
description models only as needed, exposing each via a restricted
interface tailored to specific tasks.

The richness of neurological phenomena---from clinical presentation to
anatomical and pathological investigations---poses a Catch-22:
structured annotations require comprehensive data, yet automating
analysis demands structured inputs. Current commercial medical LLMs are
often prohibitively large for local deployment; API-based solutions
present security concerns. Domain-specific language models have shown
remarkable performance on medical-specific tasks (Kraljevic et al.
2022), and the further introduction of retrieval-augmented generation
has been shown to improve results on question-answering and machine
language understanding tasks (Zakka et al. 2024). Furthermore, Soudani,
Kanoulas, and Hasibi (2024) and Gupta et al. (2024) both greatly improve
the performance of small models by incorporating RAG techniques in their
respective domains. In the field of neurology, preliminary models (Liu
et al. 2023; Pedersen et al. 2020; Moor et al. 2023) have laid the
groundwork, but none provide a compact, locally deployable solution with
RAG support.

NeuroBase will attempt to address the gap by:

\begin{itemize}
\tightlist
\item
  Creating custom datasets for RAG sources and evaluation.
\item
  Reducing model size to facilitate deployment in clinical environments
  without compromising on accuracy,
\item
  Incorporating RAG to enhance domain-specific query responses,
\item
  Delivering open-source access to neurologists and neuroscientists of
  varying technical expertise.
\end{itemize}

Furthermore, this project builds on previous work building NLP systems
for medical text, Neuradicon (Watkins et al. 2025) which is available as
an open-source code base (see Section~\ref{sec-artefacts}). User
interface dashboards have also been built alongside Neuradicon as a back
end; this neurodash software is also available open-source (also linked
in Section~\ref{sec-artefacts}).

\subsection{Methodology}\label{methodology-1}

The methodology for building NeuroBase consists of fine-tuning an
open-source Gemma-7b language model on a large corpus of
neurology-specific text and then evaluating the performance of the model
on a set of neurology-specific tasks. The datasets of interest are
described in Section~\ref{sec-datasets}, for the fine-tuning, evaluation
and RAG tasks. Fine-tuning uses the QLoRA method (Dettmers et al. 2023).
This method trains only a small subset of the total model, considerably
decreasing compute requirements for fine-tuning a model to a specific
task by updating only adaptors and attention weights. First, the model
is trained with a general neurology dataset comprised of case reports,
radiology reports and textbook data in a pre-training format. These data
consist of general text in an unstructured format. The key fine-tuning
dataset is constructed of question-answering pairs; this enables the
model to act as a domain-specific query system. The fine-tuning process
is performed using the Hugging Face Accelerate library (Jain 2022) to
enable efficient training on consumer-grade hardware.

For our RAG models, I utilise the embedding model supplied by the BGE
model (Xiao et al. 2023), which uses the FlagEmbedding architecture. I
embed our textbook corpora and NICE guidelines with BGE embeddings and
perform vector search via LangChain (Mavroudis 2024). Retrieved passages
are prepended to LLM prompts to improve answer relevance. The complete
configuration list is as follows:

\begin{itemize}
\item
  \textbf{Experimental Configurations}:

  \begin{itemize}
  \tightlist
  \item
    \emph{Gemma-base} without RAG
  \item
    \emph{Gemma-base} with RAG
  \item
    \emph{Gemma-finetuned} without RAG
  \item
    \emph{Gemma-finetuned} with RAG
  \end{itemize}
\item
  \textbf{Benchmarks}: GPT-4 serves as an external gold standard.
\end{itemize}

Along with simply training a model on a medical dataset, we must also
evaluate the efficacy of domain-specific tasks, including vocabulary and
linguistic constructions that are rare in general English text but are
common within the field of neurology. Without domain-specific evaluation
the reliability of these models is questionable, and the phenomenon of
hallucination that plagues LLM models restricts the trustworthiness of
such models in general use. We also compare the performance of our
models to the performance of GPT-4 on the same tasks, to ensure that we
are not simply training a model that is worse than the current
state-of-the-art; this is inspired by the work of (- Liu et al. 2023) on
radiology. The evaluation procedure follows the approach of Gupta et al.
(2024), which includes metrics for both question-answering and
summarisation tasks. The evaluation metrics are as follows:

\begin{itemize}
\tightlist
\item
  \textbf{Accuracy} on MCQA datasets.
\item
  \textbf{ROUGE, BLEU, METEOR} metrics on summarization and text
  similarity tasks.
\end{itemize}

The code and routines for fine-tuning and evaluation are implemented in
Python using the Hugging Face Transformers library and the Accelerate
library for efficient training on consumer-grade hardware. The model is
trained on a single NVIDIA RTX 4090 GPU, which is sufficient for
training the smaller Gemma models with QLoRA. The full training code and
deployment scripts are detailed in Section~\ref{sec-deployment}.

\subsection{Results}\label{results}

The following results summarise the performance of the fine-tuned models
on the neurology-specific datasets, comparing them to the baseline
Gemma-7b model and GPT-4. The results are presented for both
question-answering tasks and summarisation tasks, with and without
retrieval-augmented generation (RAG). The tables in this section show
the results for each model configuration on the respective datasets.
Table~\ref{tbl-qa} shows the performance of the fine-tuned models on
question-answering tasks, measured by accuracy.
Table~\ref{tbl-summarization} shows the performance of the fine-tuned
models on summarisation tasks, measured by ROUGE-L, BLEU, and METEOR
scores. The RAG label refers to the use of retrieval-augmented
generation, where the model is provided with additional context from a
knowledge base during inference. CR is the model fine-tuned on case
reports, QA is the model fine-tuned on question-answering pairs, TB is
the model fine-tuned on textbook data, and Omni is the model fine-tuned
on a combination of all three datasets. The results are compared to the
baseline Gemma-7b model and GPT-4/3.5.

\begin{longtable}[]{@{}lll@{}}
\caption{This table records the accuracy scores for each model on the
MRC question-answering dataset (MRCQA) and the textbook
question-answering dataset (TextbookQA). The models are evaluated both
with and without retrieval-augmented generation (RAG) to assess the
impact of RAG on performance. The results show that the fine-tuned
models, particularly with RAG, perform well on both datasets, although
GPT-4 remains the top performer.}\label{tbl-qa}\tabularnewline
\toprule\noalign{}
\textbf{Model Configuration} & \textbf{MRCQA} & \textbf{TextbookQA} \\
\midrule\noalign{}
\endfirsthead
\toprule\noalign{}
\textbf{Model Configuration} & \textbf{MRCQA} & \textbf{TextbookQA} \\
\midrule\noalign{}
\endhead
\bottomrule\noalign{}
\endlastfoot
\textbf{Gemma-7b} & 0.4444 & 0.855 \\
\textbf{Gemma-7b (RAG)} & 0.4444 & 0.958 \\
\textbf{Gemma-7b CR} & 0.2593 & 0.8300 \\
\textbf{Gemma-7b CR (RAG)} & 0.3827 & 0.9245 \\
\textbf{Gemma-7b Omni} & 0.3704 & 0.9655 \\
\textbf{Gemma-7b Omni (RAG)} & 0.3457 & 0.9655 \\
\textbf{Gemma-7b QA} & 0.3457 & 0.9265 \\
\textbf{Gemma-7b QA (RAG)} & 0.3827 & 0.9645 \\
\textbf{Gemma-7b TB} & 0.3827 & 0.811 \\
\textbf{Gemma-7b TB (RAG)} & 0.3456 & 0.899 \\
\textbf{GPT-4} & 0.5802 & 0.9105 \\
\textbf{GPT-4 (RAG)} & 0.6419 & 0.974 \\
\end{longtable}

\begin{longtable}[]{@{}
  >{\raggedright\arraybackslash}p{(\linewidth - 6\tabcolsep) * \real{0.4478}}
  >{\raggedright\arraybackslash}p{(\linewidth - 6\tabcolsep) * \real{0.1791}}
  >{\raggedright\arraybackslash}p{(\linewidth - 6\tabcolsep) * \real{0.1791}}
  >{\raggedright\arraybackslash}p{(\linewidth - 6\tabcolsep) * \real{0.1940}}@{}}
\caption{This table records the performance of the fine-tuned models on
summarisation tasks, measured by ROUGE-L, BLEU, and METEOR scores.
Unlike the question-answering evaluation, incorporating RAG does not
improve performance significantly, and the fine-tuned models generally
perform worse than GPT-4. This suggests that while RAG enhances
question-answering capabilities, it may not be as beneficial for
summarisation tasks in this
context.}\label{tbl-summarization}\tabularnewline
\toprule\noalign{}
\begin{minipage}[b]{\linewidth}\raggedright
\textbf{Model Configuration}
\end{minipage} & \begin{minipage}[b]{\linewidth}\raggedright
\textbf{rougeL}
\end{minipage} & \begin{minipage}[b]{\linewidth}\raggedright
\textbf{bleu}
\end{minipage} & \begin{minipage}[b]{\linewidth}\raggedright
\textbf{meteor}
\end{minipage} \\
\midrule\noalign{}
\endfirsthead
\toprule\noalign{}
\begin{minipage}[b]{\linewidth}\raggedright
\textbf{Model Configuration}
\end{minipage} & \begin{minipage}[b]{\linewidth}\raggedright
\textbf{rougeL}
\end{minipage} & \begin{minipage}[b]{\linewidth}\raggedright
\textbf{bleu}
\end{minipage} & \begin{minipage}[b]{\linewidth}\raggedright
\textbf{meteor}
\end{minipage} \\
\midrule\noalign{}
\endhead
\bottomrule\noalign{}
\endlastfoot
\textbf{Gemma-7b} & 0.09936 & 0.01858 & 0.12135 \\
\textbf{Gemma-7b CR} & 0.26158 & 0.07611 & 0.30739 \\
\textbf{Gemma-7b CR (RAG)} & 0.250 & 0.0676 & 0.2961 \\
\textbf{Gemma-7b Omni} & 0.2453 & 0.0689 & 0.2904 \\
\textbf{Gemma-7b Omni (RAG)} & 0.2300 & 0.0609 & 0.2730 \\
\textbf{Gemma-7b QA} & 0.1184 & 0.0165 & 0.15485 \\
\textbf{Gemma-7b QA (RAG)} & 0.1157 & 0.01485 & 0.15852 \\
\textbf{Gemma-7b TB} & 0.1143 & 0.01186 & 0.15968 \\
\textbf{Gemma-7b TB (RAG)} & 0.1191 & 0.0122 & 0.1637 \\
\textbf{GPT-3.5} & 0.23177 & 0.06887 & 0.25981 \\
\textbf{GPT-3.5 (RAG)} & 0.23055 & 0.06701 & 0.26198 \\
\end{longtable}

The results show that fine-tuned Gemma-7b models, especially when
combined with retrieval-augmented generation (RAG), achieve strong
performance on neurology-specific question-answering tasks, approaching
GPT-4 accuracy on both MRCPUK and textbook-derived datasets. RAG
consistently improves QA accuracy across models. However, for
summarisation tasks (measured by ROUGE-L, BLEU, and METEOR), RAG
provides little benefit, and fine-tuned models generally underperform
compared to GPT-4. Overall, RAG is most effective for question
answering, while summarisation remains challenging for smaller models.
Though the performance of the large GPT-4/3.5 models is superior, the
fine-tuned Gemma-7b models demonstrate competitive results, especially
with RAG, making them suitable for local deployment in clinical
settings.

\section{Local Neurobase Deployment}\label{sec-deployment}

The deployment of Neurobase is designed to facilitate local hosting and
interaction with the model, ensuring that sensitive medical data remains
secure and private. The architecture consists of a set of bash scripts
and Docker containers that automate the setup, hosting, and interaction
with the model and its associated services. This approach allows for
easy deployment on local machines or servers without relying on external
APIs or cloud services. The NeuroBase deployment framework comprises two
primary components---\textbf{bash\_scripts} for workflow automation and
\textbf{Docker} configurations for containerised service orchestration.
Together, these artefacts enable in-situ hosting, fine-tuning, and
evaluation of large language models (LLMs) within medical research and
clinical environments, ensuring reproducibility, modularity, and ease of
integration.

\subsection{Workflow Automation via Shell
Scripts}\label{workflow-automation-via-shell-scripts}

All core model management tasks are encapsulated in the
\texttt{bash\_scripts} directory. These scripts automate the end-to-end
life cycle of LLMs---ranging from fine-tuning domain-specific corpora to
generating predictions and evaluating performance. Key scripts include:

\begin{itemize}
\item
  \textbf{Model Evaluation}

  \begin{itemize}
  \tightlist
  \item
    \texttt{evaluations.sh} Executes evaluation routines across multiple
    datasets (e.g., question‐answering and clinical case reports) and
    aggregates performance metrics into the \texttt{metrics/} directory
    for downstream analysis.
  \end{itemize}
\item
  \textbf{API Interaction Examples}

  \begin{itemize}
  \tightlist
  \item
    \texttt{example\_curl\_request.sh} \& its variants
    (\texttt{copy.sh}, \texttt{\_embed.sh}, \texttt{\_local.sh})
    Demonstrate standardised \texttt{curl} invocations to interact with
    different endpoints: text generation, embedding, and
    retrieval-augmented generation (RAG), either on remote servers or
    locally hosted APIs.
  \item
    \texttt{example\_vllm.sh} Illustrates sending completion requests to
    a vLLM-based service, including specification of model version,
    prompt text, and generation parameters.
  \item
    \texttt{test\_openai\_question.sh} Provides a template for querying
    the OpenAI API with a medical question, requesting not only the
    generated answer but also token‐level log probabilities for
    interpretability.
  \end{itemize}
\item
  \textbf{Model Fine-Tuning and Hosting}

  \begin{itemize}
  \tightlist
  \item
    \texttt{fine\_tune\_gemma.sh} Initiates fine-tuning of the ``Gemma''
    LLM using Hugging Face Accelerate, with configurable hyperparameters
    and training datasets.
  \item
    \texttt{host\_embedder.sh}, \texttt{host\_llm.sh},
    \texttt{host\_rag\_api.sh} Launch Docker containers to serve
    embedding models (e.g., BAAI/bge-large-en-v1.5), text-generation
    LLMs (e.g., Gemma), and a composite RAG API, respectively---each
    exposing a dedicated HTTP port for standardised access.
  \end{itemize}
\item
  \textbf{Prediction and Orchestration}

  \begin{itemize}
  \tightlist
  \item
    \texttt{predicting.sh} Automates generation of model outputs on QA
    and case report datasets by invoking specified endpoint URLs and
    saving structured outputs for analysis.
  \item
    \texttt{run\_all.sh} \& \texttt{run\_all\_openai.sh} Provide
    high-level orchestration scripts that sequence local predictions
    (with and without RAG) across various fine-tuned and base models,
    including comparisons to OpenAI GPT-4--based workflows.
  \item
    \texttt{set\_keys.sh} Simplifies environment configuration by
    exporting necessary OpenAI API credentials.
  \end{itemize}
\end{itemize}

Collectively, these scripts establish a transparent, reproducible
pipeline that reduces manual overhead and facilitates systematic
experimentation.

\subsection{Containerisation with
Docker}\label{containerisation-with-docker}

To ensure consistency across development, testing, and production, the
project includes a suite of Dockerfiles, each tailored to a distinct
service or computational environment:

\begin{longtable}[]{@{}
  >{\raggedright\arraybackslash}p{(\linewidth - 2\tabcolsep) * \real{0.1274}}
  >{\raggedright\arraybackslash}p{(\linewidth - 2\tabcolsep) * \real{0.8726}}@{}}
\toprule\noalign{}
\begin{minipage}[b]{\linewidth}\raggedright
Dockerfile
\end{minipage} & \begin{minipage}[b]{\linewidth}\raggedright
Purpose
\end{minipage} \\
\midrule\noalign{}
\endhead
\bottomrule\noalign{}
\endlastfoot
\textbf{Dockerfile\_db} & Configures a persistent database service for
storing intermediate data, metadata, and results. \\
\textbf{Dockerfile\_hf} & Constructs an environment equipped with
Hugging Face libraries and dependencies for model training and
inference. \\
\textbf{Dockerfile\_peft} & Sets up tools and libraries required for
Parameter-Efficient Fine-Tuning (PEFT), enabling large-scale adaptation
with minimal computation. \\
\textbf{Dockerfile\_rag} & Builds a retrieval-augmented generation
service, integrating vector stores with LLM back ends to enrich
responses with external knowledge. \\
\textbf{Dockerfile\_tgi} & Prepares a Text Generation Inference (TGI)
server for scalable, low-latency text generation using optimised LLM run
times. \\
\end{longtable}

Each Docker configuration encapsulates all necessary dependencies---such
as specific Python versions, system libraries, and model weights---thus
promoting portability across hardware infrastructures and team
environments. Users can build and run these images via standard Docker
commands:

\begin{Shaded}
\begin{Highlighting}[]
\CommentTok{\# Example: build and run the RAG API container}
\ExtensionTok{docker}\NormalTok{ build }\AttributeTok{{-}f}\NormalTok{ docker/Dockerfile\_rag }\AttributeTok{{-}t}\NormalTok{ NeuroBase{-}rag:latest .}
\ExtensionTok{docker}\NormalTok{ run }\AttributeTok{{-}d} \AttributeTok{{-}{-}name}\NormalTok{ rag\_api }\AttributeTok{{-}p}\NormalTok{ 8082:8082 neurobase{-}rag:latest}
\end{Highlighting}
\end{Shaded}

\subsection{Integration and
Extensibility}\label{integration-and-extensibility}

By combining modular shell scripts with containerized services, the
NeuroBase deployment suite achieves:

\begin{enumerate}
\def\labelenumi{\arabic{enumi}.}
\tightlist
\item
  \textbf{Reproducibility:} Version-controlled Docker images and
  scripted workflows guarantee that experiments can be reliably rerun
  across diverse environments.
\item
  \textbf{Scalability:} Services can be distributed---or orchestrated
  with tools like Kubernetes---to accommodate increasing data volumes
  and concurrent users.
\item
  \textbf{Interoperability:} Standardised APIs (HTTP/JSON) allow
  seamless integration with electronic health record (EHR) systems, web
  applications, or downstream analytics pipelines.
\item
  \textbf{Extensibility:} New models, datasets, or evaluation metrics
  can be incorporated by extending existing scripts or adding new
  Dockerfiles without disrupting the core infrastructure.
\end{enumerate}

This architecture supports rapid iteration and deployment of LLMs in
sensitive medical contexts.

\section{Language Models for Phenotyping
Presentations}\label{sec-phenotypes}

Recent advances in natural language processing have enabled the
extraction of deep phenotypic representations from clinical text
(Baddour et al. 2024). By leveraging language models, our approach
derives a succinct, hierarchical latent representation of neurological
phenomena. The core idea is to autoencode the model's internal vector
representations of neurological descriptions, thereby capturing the
intricate interactions between clinical features. This hierarchical
structure not only reflects the complexity of neurological presentations
but also provides a data-driven basis for rich phenotyping (Lindsey et
al. 2025; Marks et al. 2024). By interrogating the latent space of our
language models, we can uncover deep phenotypic patterns that are
hierarchically organised, offering a nuanced view of neurological
presentations. These deep phenotypes extend beyond simple feature
extraction, capturing complex inter-dependencies between different
clinical characteristics. The resulting representations have significant
utility in advancing both the scientific understanding and clinical
application of automated neurology phenotyping. Such encodings are also
key for interpreting the generations produced by language models.

\subsection{Methods}\label{methods}

The first approach to identify hierarchical phenotypes is to use a
language model to encode the clinical text, and then apply clustering
techniques to the resulting latent space. This allows us to identify
clusters of terms that are closely related in the latent space of the
language model, revealing how different clinical features are
interconnected and how they contribute to the overall presentation of
neurological conditions. Term vectors can be extracted from the
embeddings of our custom language model (described in
Section~\ref{sec-model}), and then they are reduced into a
low-dimensional latent space using methods such as autoencoders (the
exact model architecture is described in Watkins et al. (2025)). The
resulting latent space is then clustered using hierarchical
agglomerative clustering (HAC) using Ward linkage (Ward Jr 1963) to
identify groups of terms that are closely related in the latent space.
This approach allows us to uncover hidden structures in the data,
revealing how different clinical features are interconnected and how
they contribute to the overall presentation of neurological conditions.

The first attempt involved ICD10 (Organization 1992) codings from our
corpus of radiological reports (Section~\ref{sec-datasets}). Each
clinical entity of interest (such as the mention of a glioblastoma) was
aligned to its corresponding ICD10 code (C71 in this example). Thus we
have a set of texts, each with several mentions of clinical entities.
Each of these is labelled with the corresponding ICD10 code. The terms
are vectorised using the language model embeddings, and we can then
create latent space representations of each ICD10 based on the terms as
they appear in the text. Once the vectors are passed through the
autoencoder, we can then apply clustering techniques to the resulting
latent space. The autoencoder dimensionality reduction is necessary to
avoid the `curse of dimensionality' which is necessary for good
performance in distance-based algorithms such as HAC. By reducing the
dimensionality, we can capture the essential features of the data while
mitigating the effects of noise and irrelevant variations.

\begin{figure}

\centering{

\pandocbounded{\includegraphics[keepaspectratio]{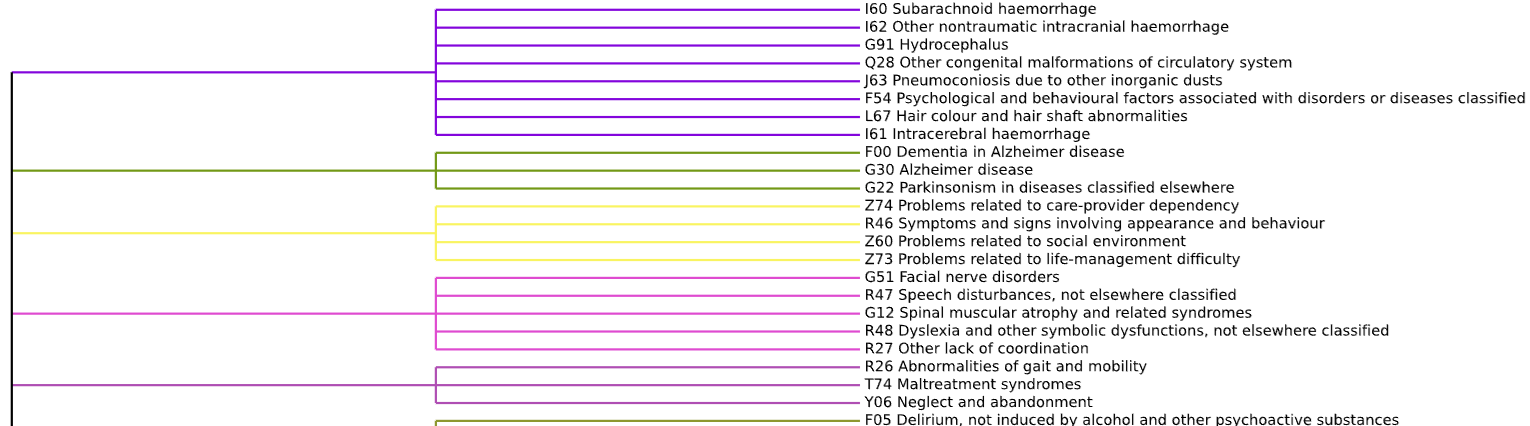}}

}

\caption{\label{fig-phenotype}One can use hierarchical phenotyping to
create a data-driven taxonomy of ICD10 codes based on their phenotype.
This figure shows a subset of the whole ICD10 tree, focusing on
neurological conditions.}

\end{figure}%

Figure~\ref{fig-phenotype} shows an example of how hierarchical
phenotyping can be used to create a data-driven taxonomy of ICD10 codes
based on their phenotype. This approach allows us to visualise the
relationships between different neurological conditions and their
associated clinical features, providing a structured framework for
understanding complex presentations. The link to the full tree can be
found in Section~\ref{sec-artefacts}

An alternative approach is the nested stochastic block model (SBM)
(Peixoto 2014). This method takes a graph (either a co-occurrence graph
or a latent-space distance graph) and approximates a nested structure of
communities within the graph. This allows us to identify groups of terms
that are closely related in the latent space of the language model,
revealing how different clinical features are interconnected and how
they contribute to the overall presentation of neurological conditions.
In this approach, we take the graph of terms directly, selected using
the multi-word expression algorithm (Section~\ref{sec-vocab}). These
term expressions are vectorised in the same manner as the HAC approach,
but instead, I use an SBM to identify a taxonomy of terms.

\begin{figure}

\centering{

\pandocbounded{\includegraphics[keepaspectratio]{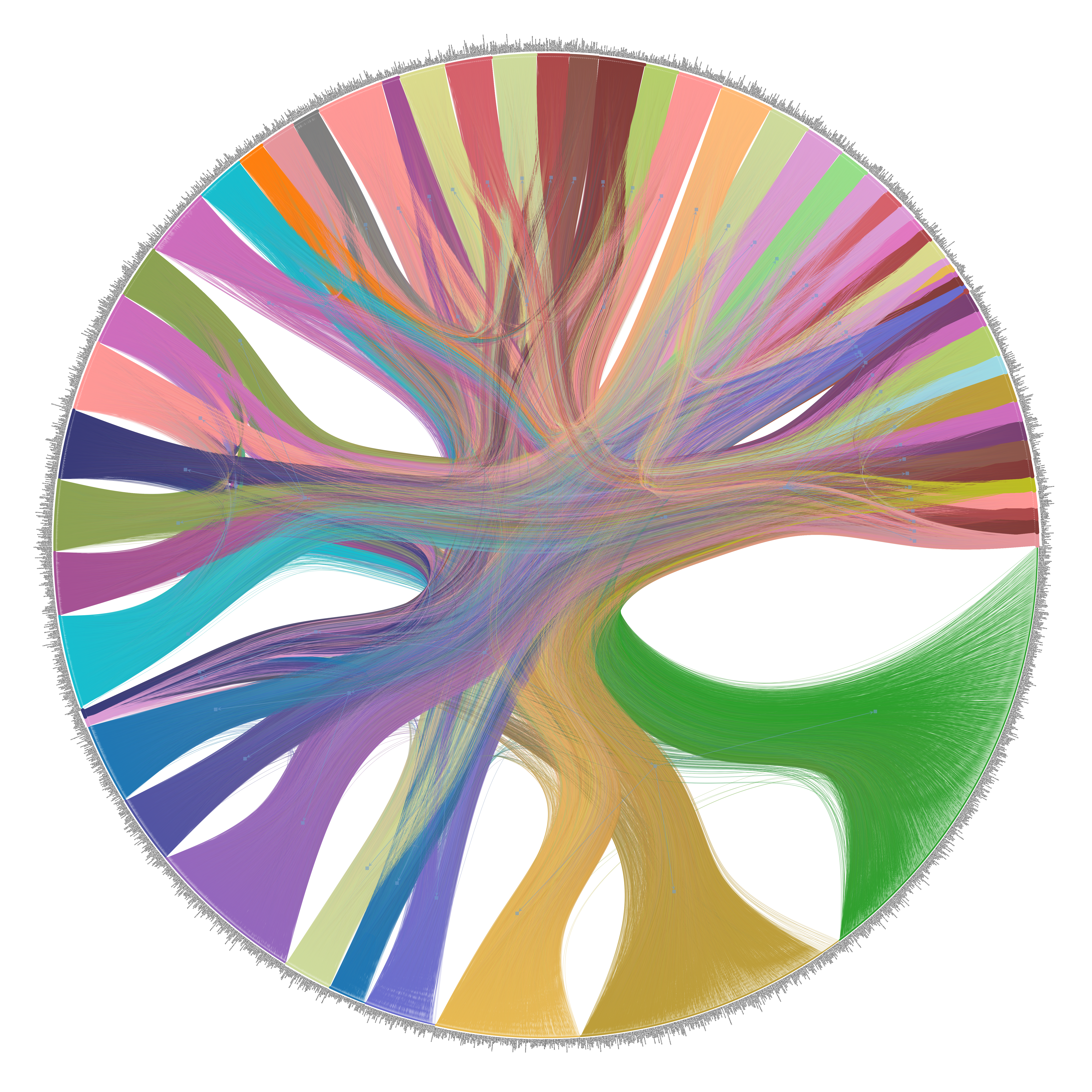}}

}

\caption{\label{fig-sbm}A stochastic block model can also be used to
identify groups of terms that are closely related in the latent space of
the language model. This approach allows us to uncover hidden structures
in the data, revealing how different clinical features are
interconnected and how they contribute to the overall presentation of
neurological conditions.}

\end{figure}%

Figure~\ref{fig-sbm} shows the resulting graph from the nested
stochastic block model. The nodes represent terms, and the edges
represent the relationships between them. The colours indicate the
different communities identified by the SBM, which correspond to
different clinical features or presentations. Each cluster or community
represents a group of terminology that is closely related both in terms
of semantics and presentation.

Tools for the representation of texts and analysis of structure in the
form of vector embeddings are part of the NeuroBase code repository (see
links in Section~\ref{sec-artefacts}). To help the exploration and
visualisation of latent spaces, I have also created the labellasso tool.
This tool enables the user to select, inspect and label latent spaces
and scatter plots for the purpose of identifying characteristic groups
or quickly labelling data for downstream models. A link to labellasso
can be found in Section~\ref{sec-artefacts}.

\section{Terminology Graphs}\label{sec-termgraph}

Understanding how neurological conditions are described in medical
literature is critical for improving communication, advancing research,
and enhancing patient care. Neurological descriptions often involve
complex, nuanced language, which can obscure underlying relationships
between conditions and treatments. By leveraging computational
approaches to analyse these descriptions, I can uncover semantic
patterns that enhance our understanding of neurological discourse. This
section builds a network of terms from neurological descriptions using
point-wise mutual information (PMI) and applies graphical modelling to
identify key semantic structures. I isolate an ontological structure
from reporting in an unsupervised manner, find key patterns of
presentation, and find the key correlates of disorders with their
description.

Unstructured text in radiological reports poses significant barriers to
effective diagnosis and clinical standardisation. Neurological
radiological reports, in particular, contain diverse terminology and
data linked to complex conditions such as strokes and epilepsy. Despite
their richness, these reports lack standardised frameworks for
terminology usage, making it difficult to extract actionable insights or
ensure consistency across clinical settings. Advances in natural
language processing, particularly transformer-based models like BERT
(Devlin et al. 2019) have shown promise in analysing medical texts. For
example, a radiology-specific variant RadBERT (Yan et al. 2022), has
demonstrated superior performance to traditional NLP models in tasks
such as abnormal sentence classification and report coding. However,
while these approaches have been applied to general radiology or
specific imaging modalities like chest X-rays, their application to
neurological radiological reports remains limited.

This section builds on prior work by integrating language model-derived
embeddings with graph-based analytical techniques (see
Section~\ref{sec-phenotypes}) to uncover latent structures within
neurological radiological reports. By identifying clusters of terms and
analysing their relationships through co-occurrence graphs, we aim to
reveal diagnostic patterns and terminology usage variations that can
inform clinical practice and standardisation efforts. In this project, I
find the distinct meta-language of neuroradiology. It aims to uncover
diagnostic patterns and terminology usage in neurology by analysing
co-occurrence graphs and clustering neurological terms. The study also
explores how these clusters correspond to specific conditions (e.g.,
strokes, epilepsy, neurodegenerative disorders) and vary across patient
demographics, providing insights into diagnostic frameworks and
terminology standardization.

In short, understanding latent structures in radiological reports can:

\begin{itemize}
\tightlist
\item
  Improve diagnostic accuracy by identifying patterns in terminology
  usage.
\item
  Standardise medical language, reducing inconsistencies across clinical
  settings.
\item
  Enhance knowledge representation for electronic health records (EHRs),
  enabling better decision-making tools for clinicians.
\end{itemize}

\subsection{Methods}\label{methods-1}

The project begins with the extraction of terms and phrases from a
corpus of neurological texts, followed by the computation of PMI to
identify statistically significant co-occurrences of terms. These
co-occurrences are represented as a weighted graph, where nodes are
terms and edges reflect their mutual information. A graph model is
trained on this graph to identify higher-order structures and patterns.
Semantic sub-networks can be extracted and analysed, focusing on the
relationships between conditions, interventions, and descriptors.

The study employs:

\begin{itemize}
\tightlist
\item
  Graph-based techniques such as co-occurrence graphs and modularity
  analysis to detect latent structures.
\item
  Clustering algorithms to group terms and analyse diagnostic patterns.
\end{itemize}

We extracted terms and phrases from a large corpus of neurological
texts---including over 300,000 MRI head scans collected during 25 years
of clinical practice at Queen Square---to capture the diverse language
used in radiological reports (see Section~\ref{sec-datasets}). To
identify statistically significant co-occurrences, I computed the
Point-wise Mutual Information (PMI) between terms and represented these
relationships as a weighted graph where nodes correspond to terms and
edges denote their mutual information. High PMI values might indicate
meaningful relationships, such as common co-morbidities or
symptom-disease relationships.

Graph-based techniques were employed to further analyse the data. I used
soft Louvain clustering (Blondel et al. 2008) to identify communities
within the network, allowing for the detection of non-exclusive thematic
groups. Unlike traditional Louvain clustering, soft Louvain clustering
ascribes a probability of membership to each node in the graph,
therefore I can use this parameter to find non-exclusive communities of
nodes in the graph. The value of non-exclusive community detection is
that it allows each node to belong to more than one community. This is
certainly the case for non-specific terminology such as ``inflammation''
that can co-occur alongside multiple different conditions or diagnoses.
These communities might correspond to thematic areas within a corpus,
such as specific branches of medicine or research clusters.
Investigating these communities can reveal how different topics are
interconnected and identify potential interdisciplinary research areas.
A threshold was applied to aggregate nodes for clarity, while
Eigenvector Centrality (Bonacich 1972) was used to pinpoint the most
influential terms within each cluster. These high-centrality primary
nodes served as the ``head'' terms in an ontology, with the associated
clusters forming templates that describe pathological conditions,
imaging presentations, and key anatomical locations.

In addition to community detection, other graph metrics such as the
clustering coefficient can be applied to the terminology graph to
uncover structure. High clustering coefficients suggest the presence of
specialised subtopics or highly focused areas within the medical texts.
These cliques could represent highly interconnected fields, such as
specific diseases with multiple associated symptoms or treatments.

\subsection{Results}\label{results-1}

The analysis reveals distinct semantic clusters corresponding to
different neurological conditions, such as oncological and degenerative
diseases. Sub-networks highlight how oncological conditions are
interrelated through shared descriptors, suggesting potential overlaps
in diagnostic and therapeutic approaches. Here I show four example
sub-graphs extracted from the larger terminology graph.

\begin{figure}

\centering{

\pandocbounded{\includegraphics[keepaspectratio]{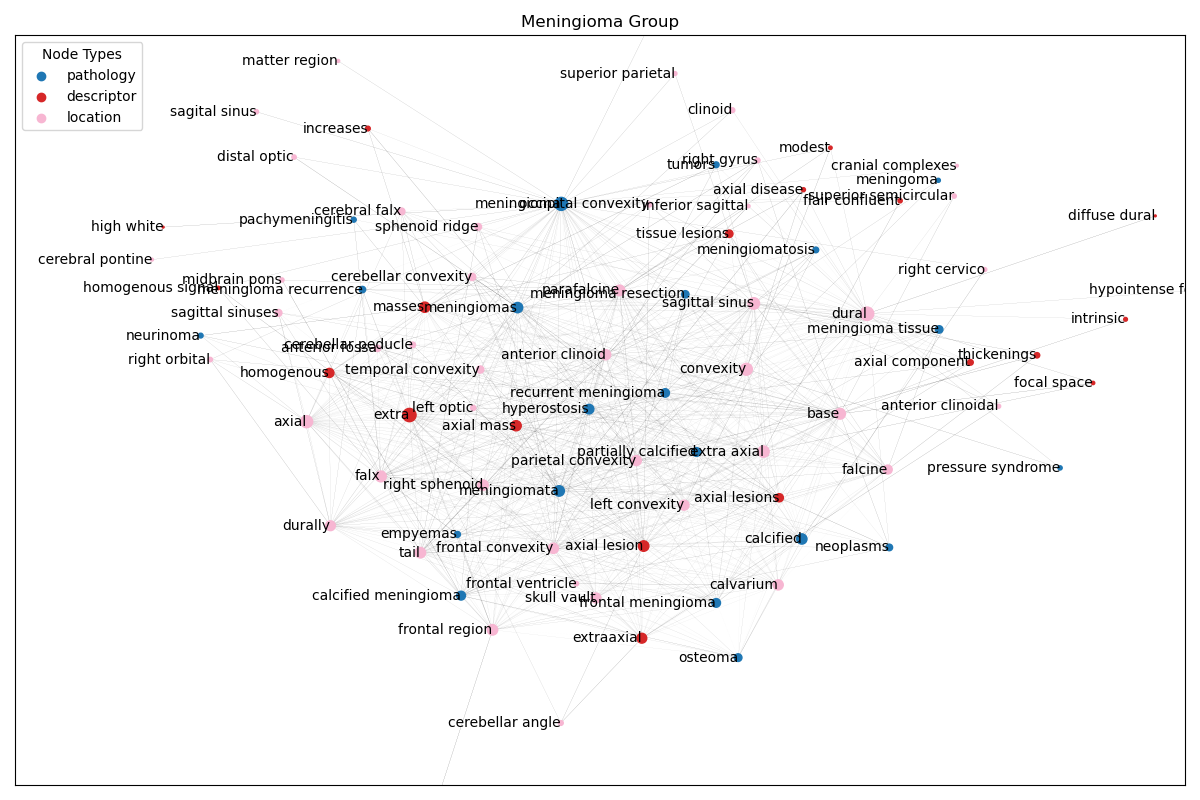}}

}

\caption{\label{fig-meningioma}Here we take the semantic sub-graph
identified with the Meningioma term. This sub-graph represents the
semantic network of terminology co-occurrent with meningiomas and their
associated description and locations.}

\end{figure}%

\begin{figure}

\centering{

\pandocbounded{\includegraphics[keepaspectratio]{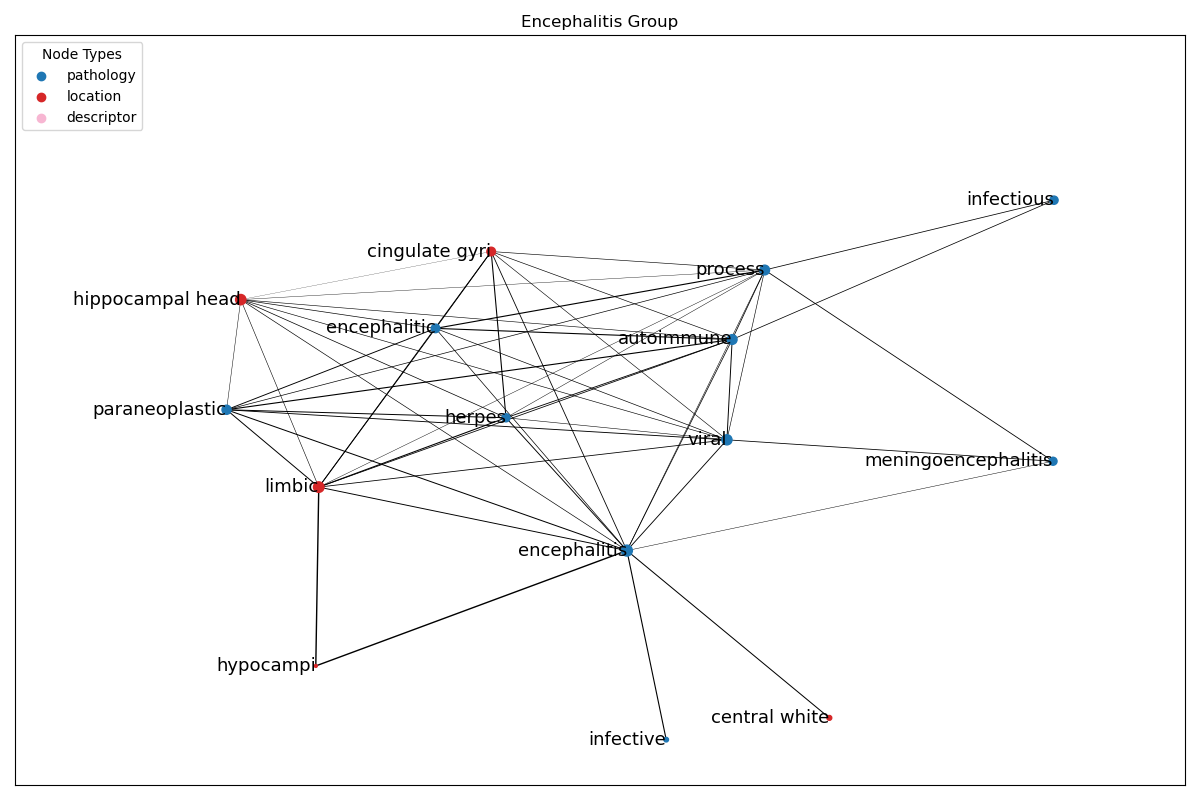}}

}

\caption{\label{fig-encephalitis}This encephalitis group shows the key
terminology that co-occurs alongside encephalitis. Note the key
locations of the limbic system and cingulate gyri, both anatomical
locations consistent with common presentations of this pathology.}

\end{figure}

Figure~\ref{fig-meningioma} shows a large semantic sub-graph identified
with the Meningioma term. The blue, red and pink nodes represent
pathological, descriptive and location terms respectively. With this
layout, we can identify the primary terminology concurrent with
meningiomas (the direct connections to meningioma terms) and also the
second-order terminology. Figure~\ref{fig-encephalitis} shows the
encephalitis group, which is a cluster of terms related to encephalitis,
a condition characterised by inflammation of the brain.

\begin{figure}

\centering{

\pandocbounded{\includegraphics[keepaspectratio]{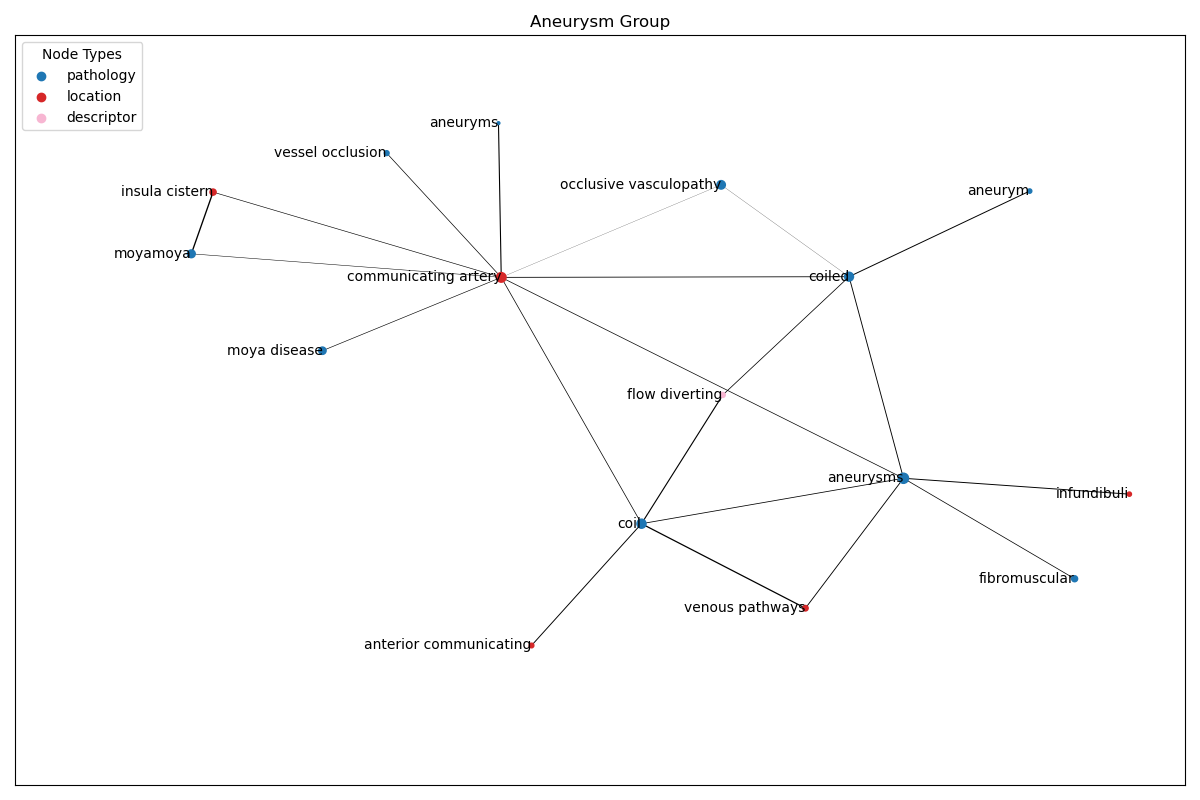}}

}

\caption{\label{fig-aneurysm}This aneurysm sub-graph shows the key
terminology associated with the presentation of aneurysms in practice.
One can see that coiling, a primary intervention used in the treatment
of aneurysms, features prominently in the network.}

\end{figure}

\begin{figure}

\centering{

\pandocbounded{\includegraphics[keepaspectratio]{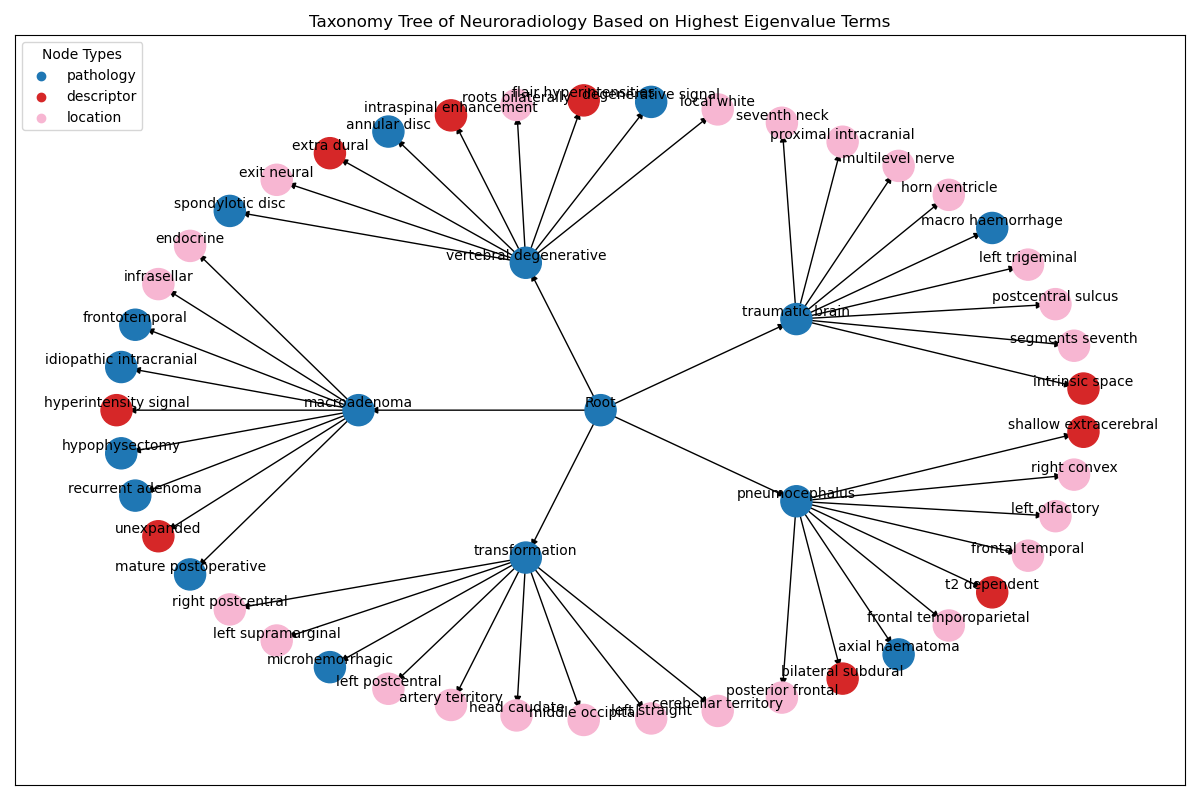}}

}

\caption{\label{fig-taxonomy}The clustering and graphical structure
allow us to create an automated ontology based on the structure and
centrality terms. Here a subset of the clusters, identified by their
most important term, shows the sub-nodes that exemplify their
presentation.}

\end{figure}

Figure~\ref{fig-aneurysm} shows the aneurysm sub-graph. This sub-graph
is smaller and was extracted using a higher probability threshold in the
Louvain community detection routine. As such, this group represents a
much tighter terminology set associated with aneurysms.
Figure~\ref{fig-taxonomy} shows a small subset of the whole ontology
created by identifying the high-centrality terms (which define
sub-specialities) and their constituent terminology that defines their
wider presentation. This method provides an entirely data-driven
approach to creating ontologies directly from medical text, without the
need for manual curation or a-priori input.

\subsection{Discussion}\label{discussion}

This section demonstrates the power of combining PMI-based networks and
graph methods to analyse medical text. By identifying key semantic
structures in neurological descriptions, the approach provides valuable
insights into medical language that can improve patient care and support
research. The findings suggest that similar methods could be applied to
other domains of medical and scientific text, bridging the gap between
unstructured data and actionable knowledge. With this approach, I can
(1) identify clusters of related terms and provide insights into
diagnostic patterns that can aid clinicians in recognising complex
conditions more effectively. (2) The observed inconsistencies underscore
the need for standardised medical language across clinical settings.
Standardization could improve communication among healthcare
professionals and reduce diagnostic variability. (3) Insights from
co-occurrence graphs can inform better knowledge representation within
EHR systems by highlighting key terms and their relationships. (4)
Variations in terminology usage across demographics suggest potential
biases that should be addressed to ensure equitable care. (5) Expanding
this approach to other medical domains or integrating multimodal data
(e.g., imaging features) could further enhance its applicability.
Additionally, exploring advanced graph neural networks could provide
deeper insights into term interconnections.

\section{Safety, Alignment, and Security in
NeuroBase}\label{sec-alignment}

This section builds on the AI safety, alignment, and security principles
outlined in the Turing Institute AI Safety Document (Leslie 2019). I
translate those high-level recommendations into concrete practices for
NeuroBase, our neurology-focused foundation model, ensuring both
domain-specific performance and adherence to rigorous clinical ethics
and governance standards. The aim was to create a guide rail based on
these principles that would determine the output of any user
interaction/query. This work package was the last part of the project,
and so only an outline has been completed.

The following sections detail our approach to embedding safety,
alignment, and security into NeuroBase's architecture, development life
cycle, and operational practices. The first process requires embedding
safety into the technical design from the outset when writing the
software. The second is a detailed plan for governance, and the third is
an alignment process that will be used to ensure the model remains safe
and aligned with clinical needs throughout its operational lifetime.

As a query-based text system, NeuroBase will implement these safety
measures at the point of user interaction. The system will pass each
query and result through an intermediary system, ensuring that all
outputs are vetted against the safety and alignment criteria outlined
below. This intermediary will serve as a filter, applying the necessary
checks before any information is presented to the user.

\subsection{Technical Safety
Objectives}\label{technical-safety-objectives}

NeuroBase must meet four intertwined operational
goals---\textbf{accuracy}, \textbf{reliability}, \textbf{security}, and
\textbf{robustness}---across all stages of model development,
deployment, and maintenance:

\begin{itemize}
\tightlist
\item
  \textbf{Accuracy \& Performance Metrics}

  \begin{itemize}
  \tightlist
  \item
    Define neurology-specific KPIs in collaboration with clinical
    stakeholders: accuracy on MRCPUK and textbook MCQA (target > 85\%),
    BLEU/ROUGE/METEOR for case-report summarisation (target BLEU > 0.25,
    ROUGE-L > 0.3).
  \item
    Continuously monitor performance against these thresholds in staging
    environments; trigger human review when metrics fall below clinical
    risk tolerances.
  \item
    Incorporate confidence estimation (e.g., calibration curves, entropy
    measures) so that low-confidence outputs can be flagged for expert
    oversight.
  \end{itemize}
\item
  \textbf{Reliability}

  \begin{itemize}
  \tightlist
  \item
    Establish deterministic inference pipelines by fixing random seeds,
    containerising environments, and versioning all dependencies.
  \item
    Implement continuous integration (CI) tests that run a suite of
    canonical neurology cases (e.g., classic Guillain--Barré syndrome
    vignette) to verify reproducible outputs under fixed retrieval
    contexts.
  \item
    Use canary deployments to gradually roll out model updates to
    controlled user groups, comparing outputs against the baseline.
  \end{itemize}
\item
  \textbf{Security}

  \begin{itemize}
  \tightlist
  \item
    Protect data integrity by:

    \begin{itemize}
    \tightlist
    \item
      Sanitising inputs to prevent code injection.
    \item
      Encrypting data for all PDF ingestion, embedding storage, and
      inference requests.
    \end{itemize}
  \item
    Secure model artefacts and endpoints:

    \begin{itemize}
    \tightlist
    \item
      Store fine-tuned adaptors and embedding indexes in
      access-controlled vaults.
    \item
      Enforce RBAC on inference APIs: only authorized clinical
      applications can query NeuroBase.
    \end{itemize}
  \end{itemize}
\item
  \textbf{Robustness}

  \begin{itemize}
  \tightlist
  \item
    Conduct systematic stress testing with:

    \begin{itemize}
    \tightlist
    \item
      Out-of-distribution texts (e.g., rare metabolic disorders
      presenting neurologically).
    \item
      Adversarially perturbed inputs (e.g., misspellings, anonymised
      identifiers) to evaluate tolerance.
    \end{itemize}
  \item
    Automate drift detection by monitoring embedding similarity metrics
    over live data streams; schedule retraining or index refresh when
    cosine similarity distributions shift beyond preset thresholds.
  \end{itemize}
\end{itemize}

\subsection{Alignment and Governance}\label{alignment-and-governance}

Adopting an \textbf{Ethical Platform} approach, we embed transparency,
accountability, and auditability into NeuroBase's life cycle:

\begin{enumerate}
\def\labelenumi{\arabic{enumi}.}
\tightlist
\item
  \textbf{Process-Based Governance (PBG) Framework}

  \begin{itemize}
  \tightlist
  \item
    Maintain a comprehensive \textbf{Dataset Factsheet} that logs:

    \begin{itemize}
    \tightlist
    \item
      Source provenance (e.g., journal name, textbook edition).
    \item
      Preprocessing steps (GROBID version, TEI-to-JSON conversion
      parameters).
    \item
      RAG index creation timestamps and embedding model versions.
    \end{itemize}
  \item
    Version-control all pipeline artefacts (data transforms, adaptor
    weights) in a secure registry for reproducibility and audit.
  \end{itemize}
\item
  \textbf{Accountability-by-Design}

  \begin{itemize}
  \tightlist
  \item
    Assign domain experts as pipeline stewards: one neurologist oversees
    dataset curation; one ML engineer owns fine-tuning; one DevOps lead
    handles deployment.
  \item
    Log every inference transaction with anonymised input metadata and
    output summaries; store logs for at least two years to support
    retrospective analysis.
  \end{itemize}
\item
  \textbf{Transparency \& Explainability}

  \begin{itemize}
  \tightlist
  \item
    Surface RAG provenance in inference responses:

    \begin{itemize}
    \tightlist
    \item
      Attach document identifiers, section headers, and retrieval
      relevance scores alongside model answers.
    \item
      Provide an optional ``explanation mode'' where underlying
      token-level attributions (e.g., attention heatmaps) annotate the
      generated text.
    \end{itemize}
  \end{itemize}
\end{enumerate}

By integrating these detailed safety, alignment, and security practices
into NeuroBase's modular RAG-enabled and fine-tuned architecture, we
ensure that our foundation model not only achieves cutting-edge
neurology performance but also upholds the highest standards of clinical
trustworthiness, data security, and ethical accountability.d

\section{Discussion \& Future Work}\label{sec-discussion}

This fellowship set out to build a bespoke ``Neurobase'' language model,
alongside associated software and tooling to help neurologists and
neuroscientists interact with a domain-specific large language model,
but the project evolved in response to recent advances. The rapid
evolution of large language models and retrieval‐augmented generation
paradigms has fundamentally reshaped the landscape for domain-specific
NLP applications. Our initial aim---to fine-tune a bespoke foundational
model (``Neurobase'') exclusively on in-house neurological texts---was
overtaken by the emergence of broadly pre-trained medical LLMs
exhibiting comparable or superior performance when paired with targeted
retrieval mechanisms. Empirical evaluations on three newly introduced
neurology datasets (MRCPUK exam questions, clinical case reports, and
textbook‐derived MCQA) demonstrated that small, adaptor-based models
(Gemma+QLoRA) augmented with RAG nearly matched GPT-4's accuracy on
diagnostic MCQA tasks, while significantly reducing compute and
deployment overhead.

Beyond model performance, our Dockerized deployment framework and
shell-script orchestration ensure that Neurobase components
(fine-tuning, embedding services, RAG API) can be hosted securely within
hospital firewalls, preserving patient confidentiality. Initial safety
and alignment measures---drawing on FAST-track principles---demonstrated
the feasibility of embedding technical safeguards (input sanitisation,
drift detection, canary testing) and governance protocols (dataset
factsheets, inference logging) within a research-grade LLM pipeline.
However, comprehensive clinical validation and integration into
electronic health records remain future work.

NeuroBase uniquely combines small-model efficiency with RAG-enhanced
performance for neurology, delivering:

\begin{itemize}
\tightlist
\item
  \textbf{Local Deployability}: Eliminates reliance on external APIs,
  ensuring patient data security.
\item
  \textbf{Domain Fidelity}: Tailored evaluation on neurology-speciality
  datasets mitigates rare-terminology omissions.
\item
  \textbf{Accessibility}: Open-source release empowers researchers
  without extensive ML infrastructure.
\end{itemize}

In addition to language modelling, this project also explored using
point's-mutual-information graphs and nested SBM clustering to surface
clinically intuitive communities (e.g.~meningioma--dural tail--falx,
limbic encephalitis--cingulate--mesial temporal lobes). Mapping those to
ICD-10 codes or imaging phenotypes produces the beginnings of a
data-driven ontology---a foundation for explainable triage tools,
decision support and audit dashboards. Early experiments with
hierarchical HAC corroborated the idea that LLM embeddings encode latent
``deep phenotypes''. The multi-word expression (MWE) toolkit pyMWE
proved effective in extracting high-value bigrams (e.g.~``small vessel
ischaemia'', ``left frontal lesion''), laying the groundwork for
downstream vocabulary enrichment and improving retrieval relevance in
RAG pipelines. These data-driven insights not only validate known
clinical associations (such as coiling as the primary aneurysm
intervention) but also surface subtler term relationships that merit
further clinical investigation.

\textbf{Future Directions} include:

\begin{enumerate}
\def\labelenumi{\arabic{enumi}.}
\tightlist
\item
  \textbf{Multimodal Integration}: Adapting open-source multimodal
  architectures (e.g.~phi-4-multimodal) to jointly process radiological
  images and text, enabling richer phenotypic representations.
\item
  \textbf{Longitudinal Phenotyping}: Leveraging deep latent
  representations to model temporal trajectories of neurological
  disease, linking text-derived phenotypes to imaging bio-markers and
  outcomes data.
\item
  \textbf{Interactive Clinical Tools}: Building user-friendly interfaces
  (e.g.~Shiny apps or web dashboards) enabling neurologists to query the
  ontology graph, refine MWE thresholds, and inspect RAG provenance in
  real-time.
\item
  \textbf{Regulatory Alignment}: Collaborating with clinical governance
  bodies to validate model outputs against established guidelines, and
  to navigate data protection regulations (e.g.~GDPR, NHS IG Toolkit).
\end{enumerate}

This project demonstrates that a hybrid approach---marrying small,
locally deployable language models with retrieval-augmented generation
and graph-based semantic analyses can achieve high performance on
neurology-specific tasks while maintaining data security and operational
feasibility. The creation of three novel neurology datasets, the
open-source pyMWE, labellasso and NeuroBase tool-kits, and the neuro
language ontology graphs collectively advance the state of the art in
domain-specific NLP for neurology. Our modular, containerized
architecture ensures portability across clinical environments, and our
preliminary safety and alignment measures lay the groundwork for
responsible deployment. By shifting focus from bespoke foundational
models to adaptable RAG pipelines, we provide a framework that can
extract structured, actionable insights from unstructured clinical text.
As multimodal and longitudinal extensions are developed, this work will
continue to bridge the gap between free-text neurological records and
precision clinical decision support.

\section{Project Artefacts}\label{sec-artefacts}

This section lists all the primary project artefacts such as datasets
and code bases and means of accessing them.

Datasets produced during this project can be accessed upon request but
may only be used within the institution. These datasets are in both
Huggingface and raw formats. The datasets are:

\begin{itemize}
\tightlist
\item
  \textbf{Case report pre-training/RAG} dataset
\item
  \textbf{Case report summarisation} dataset
\item
  \textbf{Textbook pre-training/RAG} dataset
\item
  \textbf{TextbookQA} dataset
\item
  \textbf{Radiology reports} dataset. Note this particular dataset is
  not available for public use.
\item
  \textbf{NICE guidelines} dataset
\item
  The speciality exam \textbf{MRCPUK} dataset
\end{itemize}

Software

\begin{itemize}
\tightlist
\item
  The \textbf{pyMWE} software can be found at its
  \href{https://github.com/henrywatkins/pymwe}{github repo}
\item
  The \textbf{NeuroBase} deployment and training repository can be found
  at its \href{https://github.com/henrywatkins/neurobase}{github repo}
\item
  Extra code for the \textbf{terminology graph}, and the ICD10 code
  dendrogram can be found at its
  \href{https://github.com/henrywatkins/neuro-term-graph}{github repo}
\item
  \textbf{Neuradicon} can be found via its
  \href{https://github.com/high-dimensional/neuradicon}{github repo}
\item
  \textbf{labellasso} can be accessed via its
  \href{https://github.com/henrywatkins/labellasso}{github repo}
\item
  The \textbf{neurodash} dashboard software is available at the
  \href{https://github.com/high-dimensional/neurodash}{neurodash github
  repo}
\end{itemize}

\phantomsection\label{refs}

\end{document}